\documentclass{article}

    \usepackage[final,nonatbib]{neurips_2020}

\usepackage[utf8]{inputenc} 
\usepackage[T1]{fontenc}    
\usepackage{hyperref}       
\usepackage{url}            
\usepackage{booktabs}       
\usepackage{amsfonts}       
\usepackage{nicefrac}       
\usepackage{microtype}      

\title{Privacy Preserving Demand Forecasting to Encourage Consumer Acceptance of Smart Energy Meters}

\author{%
  Christopher Briggs\\
  School of Computing \\
  \& Mathematics\\
  Keele University\\
  Keele, Staffordshire, UK\\
  \texttt{c.briggs@keele.ac.uk} \\
  \And
  Zhong Fan\\
  School of Computing \\
  \& Mathematics\\
  Keele University\\
  Keele, Staffordshire, UK\\
  \texttt{z.fan@keele.ac.uk} \\
  \And
  Peter Andras\\
  School of Computing \\
  \& Mathematics\\
  Keele University\\
  Keele, Staffordshire, UK\\
  \texttt{p.andras@keele.ac.uk} \\
}

\begin{document}

\maketitle

\begin{abstract}
  In this proposal paper we highlight the need for privacy preserving energy demand forecasting to allay a major concern consumers have about smart meter installations. High resolution smart meter data can expose many private aspects of a consumer's household such as occupancy, habits and individual appliance usage. Yet smart metering infrastructure has the potential to vastly reduce carbon emissions from the energy sector through improved operating efficiencies. We propose the application of a distributed machine learning setting known as federated learning for energy demand forecasting at various scales to make load prediction possible whilst retaining the privacy of consumers' raw energy consumption data.
\end{abstract}

\section{Motivation \& Impact}

Smart meters are being deployed in many countries across the world for the purpose of optimising efficiency within electricity grids and giving consumers insight into their energy usage. The meters record energy consumption within a building directly from the electricity supply and periodically communicate this data to energy suppliers and other entities in the energy sector. Smart meter data contain an enormous amount of potential predictive power that will aid the transition from fossil fuel technologies to cleaner and renewable technologies \cite{Anonymous:2019up}. However this high-resolution data is particularly sensitive as it can easily enable inference about household occupancy, lifestyle habits or even what and when specific appliances are being used in a household.

A large contribution of renewables in the energy mix poses a significant challenge for balancing supply and demand. If peak demand coincides with low wind/solar inputs, energy must be provided by reliable backup generation, such as idling gas turbines. Such solutions are very costly, both economically and environmentally and serve to discourage the installation of large amounts of renewable energy production. Reliable forecasting will provide opportunity for more efficient optimisation of electricity grids to cope with varying energy demand.

With a rapidly evolving global climate and swinging extremes of temperature, large heating and cooling loads will become a more significant factor in energy demand profiles. Additionally, as other areas of our daily lives become decarbonised, loads from electric vehicles etc. will pose a strain on electricity grids. Accurate forecasting is important here to understand how demand is evolving with consumer behaviour change. Demand side response applications - those that actively or suggestively seek to reduce peak loads when necessary - can also benefit from accurate realtime forecasting.

Despite the benefits for promoting a greener energy sector, smart meter installation in most countries is an opt-in process and levels of adoption of smart meters remains low. Data privacy and security concerns are among the most cited reasons consumers give for rejecting a smart meter installation \cite{BaltaOzkan:2014dy}. The reasons consumers give as privacy concerns with smart meters include government surveillance, energy companies selling data and illegal data acquisition/use \cite{McKenna:2012ga}. Although smart meter privacy has been recognised as an important issue by researchers in both energy and computing for over a decade \cite{Efthymiou:dd, 2014ISysJ...8..641K}, tackling it from a privacy preserving machine learning perspective is a novel area of research.

In this paper, we propose the use of a modern distributed machine learning setting known as federated learning to train accurate energy demand forecasting models that preserve the privacy of consumer energy consumption data to enable greater adoption of smart meters by privacy-concious consumers.

\section{Traditional Energy Demand Forecasting Approaches}

Various models, both simple and complex have been applied to energy demand forecasting. Simple methods based on autoregressive (AR) and moving average (MA) models apply a mathematical operation to demand data formatted as a time series. These models assume the time series to be stationary and work well for producing short-term forecasts, yet poorly in the longer-term as they only account for recent past historical data. Regression methods such as linear regression model the relationship between input data and a target output. These models are simple to construct and have some level of interpretability. Artificial neural networks improve on these regression models as they transform the inputs in a non-linear fashion to make predictions. The most successful neural network architectures for forecasting are based on recurrent neural networks (RNNs), such as Long-Short Term Memory networks (LSTMs). These architectures can learn what long and short term information to pay attention to during the training process. Recent surveys compare and contrast traditional and modern approaches to energy consumption forecasting and conclude that AI-based methods (such as those that utilise neural networks) offer the greatest predictive performance across all forecasting horizons \cite{Wei:2019ez,Bourdeau:2019fj}. 

The key drawback to how all these methods have been applied in the energy demand forecasting literature at present is the need for data to be centralised. Clearly the privacy of consumer energy consumption data can easily be violated in such cases.

\section{Proposed Privacy Preserving Energy Demand Forecasting Approach}
Federated learning \cite{mcmahan2017communication} is a setting that allows for distributing the training of a machine learning model over many devices or clients without the necessity for clients to share their raw datasets. In a typical federated learning setup, a central entity orchestrates the training schedule. The central entity provides a parameter server that stores a global model which the clients can read and write to. At the start of training, the parameter server is randomly initialised with its global model. In a round of training, a fraction of clients are selected to download the global model. These clients proceed to train a local model initialised from the global model on their local datasets for a pre-determined number of epochs. The difference between the global and trained local model forms a weight vector which becomes the payload that each client sends back to the parameter server. These vectors are aggregated at the parameter server, for example by taking the average of all the local models as in the FedAvg algorithm \cite{mcmahan2017communication}. This aggregated model then becomes the new global model. Training proceeds in rounds, selecting clients and training in a distributed manner until the model converges to some fixed accuracy or other desirable metric. In this setting, no raw data from any client is shared with the central entity or with other clients, affording clients some level of privacy over their data. To further protect privacy, federated learning can be combined with differential privacy in order to add noise to the local model updates and obscure individual contributions to the global model \cite{2017arXiv171006963B}. Additionally secure aggregation can be applied to deny the central entity the opportunity to observe individual client models at all, only in aggregate \cite{Bonawitz:2017gu}.

For the purpose of forecasting, any model that can be iteratively optimised to minimise an objective function of the form:

\begin{equation}
  \label{eq:risk-minimisation}
  \min_{w \in \mathbb{R}^d} \frac{1}{m} \sum_{i=1}^{m} f_i(w).
\end{equation}

can be trained under the federated learning setting. Here $w$ is a parameterised model, $f_i$ is a local objective function and $m$ is the number of clients over which the aggregation is taking place. This covers many forecasting algorithms from simple linear regressions to very complex recurrent neural networks. LSTM's have been shown to have exceptional predictive power where long and short-term dependencies are observed in a dataset \cite{Kong:cn}. This is the case in energy consumption data where seasonal cycles, local weather conditions, day of the week and time of day affect the energy consumption of individual households.

For our purposes, a high quality dataset is publicly available from the Low Carbon London project \cite{Anonymous:2015wr} that was conducted between 2011 and 2014, led by UK Power Networks. This dataset contains half-hourly energy consumption readings for 5,567 households in London, UK. As the households are all in close proximity, spread across a single city, the data can be combined with historic weather data such as temperature and humidity. This weather data adds valuable signals into the forecasting model to improve accuracy as is applied in most of the energy demand forecasting literature.

To evaluate the effectiveness of federated learning in this domain, we plan to contrast this approach with traditional learning using the same model architecture on centralised data which is typical of all previous energy demand forecasting research. We hope to show that short-term, medium-term and long-term forecasting models can compete with centralised learning whilst also preserving the privacy of consumers' energy consumption data. Model accuracy is the key measurement by which to compare models, however in the forecasting literature there are several methods for measuring accuracy with varying benefits and drawbacks \cite{Bourdeau:2019fj}. The most robust metrics are the mean absolute percentage error (MAPE) and the co-efficient of variation of the root mean squared error (CV-RMSE). It is these metrics we will apply in our study in order to compare with other results in the literature. By deploying accurate forecasting in a privacy preserving manner, we expect greater acceptance from consumers to install a smart meter that will help to drive the decarbonisation of energy grids around the world.

Where forecasting is required on the supply side, secure aggregation \cite{Bonawitz:2017gu} and/or differential privacy \cite{2017arXiv171006963B} can be employed to communicate predictions from individual clients to a central entity in a privacy preserving manner. The parameters of differential privacy can be adapted to afford a lesser or greater amount of privacy which has an effect on model performance which can be quantified and measured.

\section{Challenges: Applying Federated Learning to Energy Demand Forecasting}

Federated learning adds a layer of complexity to machine learning tasks. Here we discuss some of the most significant barriers to deployment of this technology for real-world energy demand forecasting using smart meter data.

\paragraph{Communicating large models between rounds of training}
In some domains communicating model updates consumes less bandwidth than communicating the raw data itself (e.g video feeds). However for energy demand forecasting, communicating a large weight vector representing a model update vs half-hourly consumption readings represents a significant increase in bandwidth to train a model using federated learning. Additionally, smart metering infrastructure tends to use slower 2G mobile networks for communication. However methods for compressing model updates through sparsification have shown great promise to reduce the size of communication payloads with little harm to model performance \cite{Konecny:2016ts}. Reducing the fraction of clients selected to participate in training would also reduce how often individual clients need to communicate.

\paragraph{Lack of compute on current smart meters}
Smart meters deployed today have very little computational capability. As such, training a mid/large sized machine learning model would be infeasible. However, smart meters are now starting to be deployed with in-home digital displays (IHDs) which have significantly faster hardware and higher capacity to run more complex tasks such as machine learning. Moreover, IHDs can be connected to a home-based WiFi network that could aid in communicating the large weight vectors inherent in federated learning.

\paragraph{Statistical heterogeneity among participating clients}
Federated learning has been shown to perform considerably worse under non-IID assumptions, where the data distribution among clients is significantly skewed. We propose to tackle this problem by applying our previous work \cite{Briggs:it} which identifies clusters of clients that provide similar model updates and isolates further federated optimisation within each identified cluster.

\section*{Acknowledgements}
This work is partly supported by the SEND project (grant ref. 32R16P00706) funded by ERDF and BEIS. We would also like to thank Hildebrand Technology Limited for helpful discussions on the feasibility of deployment of a federated learning energy demand forecasting application.

\bibliographystyle{IEEEtran}
\bibliography{../main}

\begin{thebibliography}{10}
\providecommand{\url}[1]{#1}
\csname url@samestyle\endcsname
\providecommand{\newblock}{\relax}
\providecommand{\bibinfo}[2]{#2}
\providecommand{\BIBentrySTDinterwordspacing}{\spaceskip=0pt\relax}
\providecommand{\BIBentryALTinterwordstretchfactor}{4}
\providecommand{\BIBentryALTinterwordspacing}{\spaceskip=\fontdimen2\font plus
\BIBentryALTinterwordstretchfactor\fontdimen3\font minus
  \fontdimen4\font\relax}
\providecommand{\BIBforeignlanguage}[2]{{%
\expandafter\ifx\csname l@#1\endcsname\relax
\typeout{** WARNING: IEEEtran.bst: No hyphenation pattern has been}%
\typeout{** loaded for the language `#1'. Using the pattern for}%
\typeout{** the default language instead.}%
\else
\language=\csname l@#1\endcsname
\fi
#2}}
\providecommand{\BIBdecl}{\relax}
\BIBdecl

\bibitem{Anonymous:2019up}
\BIBentryALTinterwordspacing
{Smart meter benefits: Role of smart meters in responding to climate change}.
  [Online]. Available:
  \url{https://www.smartenergygb.org/en/-/media/SmartEnergy/essential-documents/press-resources/Documents/Smart-Energy-GB-report-2---Role-in-climate-change-mitigation-Final_updated-300819.ashx?la=en&hash=E353C643DBFA500BDA2DDBB55F599575B1FE21B5}
\BIBentrySTDinterwordspacing

\bibitem{BaltaOzkan:2014dy}
N.~Balta-Ozkan, O.~Amerighi, and B.~Boteler, ``{A comparison of consumer
  perceptions towards smart homes in the UK, Germany and Italy: reflections for
  policy and future research},'' \emph{Technology Analysis {\&} Strategic
  Management}, vol.~26, no.~10, pp. 1176--1195, Dec. 2014.

\bibitem{McKenna:2012ga}
E.~McKenna, I.~Richardson, and M.~Thomson, ``{Smart meter data: Balancing
  consumer privacy concerns with legitimate applications},'' \emph{Energy
  Policy}, vol.~41, pp. 807--814, Feb. 2012.

\bibitem{Efthymiou:dd}
C.~Efthymiou and G.~Kalogridis, ``{Smart Grid Privacy via Anonymization of
  Smart Metering Data},'' in \emph{2010 1st IEEE International Conference on
  Smart Grid Communications (SmartGridComm)}.\hskip 1em plus 0.5em minus
  0.4em\relax IEEE, pp. 238--243.

\bibitem{2014ISysJ...8..641K}
G.~Kalogridis, M.~Sooriyabandara, Z.~Fan, and M.~A. Mustafa, ``{Toward Unified
  Security and Privacy Protection for Smart Meter Networks},'' \emph{IEEE
  Systems Journal}, vol.~8, no.~2, pp. 641--654, Jun. 2014.

\bibitem{Wei:2019ez}
N.~Wei, C.~Li, X.~Peng, F.~Zeng, and X.~Lu, ``{Conventional models and
  artificial intelligence-based models for energy consumption forecasting: A
  review},'' \emph{Journal of Petroleum Science and Engineering}, vol. 181,
  Oct. 2019.

\bibitem{Bourdeau:2019fj}
M.~Bourdeau, X.~Q. Zhai, E.~Nefzaoui, X.~Guo, and P.~Chatellier, ``{Modeling
  and forecasting building energy consumption: A review of data-driven
  techniques},'' \emph{Sustainable Cities and Society}, vol.~48, Jul. 2019.

\bibitem{mcmahan2017communication}
B.~McMahan, E.~Moore, D.~Ramage, S.~Hampson, and B.~A. y~Arcas,
  ``{Communication-Efficient Learning of Deep Networks from Decentralized
  Data},'' in \emph{Artificial Intelligence and Statistics}, 2017, pp.
  1273--1282.

\bibitem{2017arXiv171006963B}
H.~B. McMahan, D.~Ramage, K.~Talwar, and L.~Zhang, ``{Learning Differentially
  Private Recurrent Language Models},'' \emph{ICLR}, 2018.

\bibitem{Bonawitz:2017gu}
K.~Bonawitz, V.~Ivanov, B.~Kreuter, A.~Marcedone, H.~B. McMahan, S.~Patel,
  D.~Ramage, A.~Segal, and K.~Seth, ``{Practical Secure Aggregation for
  Privacy-Preserving Machine Learning},'' in \emph{Proceedings of the ACM
  SIGSAC Conference on Computer and Communications Security}.\hskip 1em plus
  0.5em minus 0.4em\relax ACM, Oct. 2017, pp. 1175--1191.

\bibitem{Kong:cn}
W.~Kong, Z.~Y. Dong, Y.~Jia, D.~J. Hill, Y.~Xu, and Y.~Zhang, ``{Short-Term
  Residential Load Forecasting Based on LSTM Recurrent Neural Network},''
  \emph{IEEE Transactions on Smart Grid}, vol.~10, no.~1, pp. 841--851, Jan.
  2019.

\bibitem{Anonymous:2015wr}
\BIBentryALTinterwordspacing
{SmartMeter Energy Consumption Data in London Households}. [Online]. Available:
  \url{https://data.london.gov.uk/dataset/smartmeter-energy-use-data-in-london-households}
\BIBentrySTDinterwordspacing

\bibitem{Konecny:2016ts}
J.~Kone{\v c}n{\'{y}}, H.~B. McMahan, F.~X. Yu, P.~Richt{\'a}rik, A.~T. Suresh,
  and D.~Bacon, ``{Federated Learning: Strategies for Improving Communication
  Efficiency},'' \emph{NIPS Workshop on Private Multi-Party Machine Learning},
  2016.

\bibitem{Briggs:it}
C.~Briggs, Z.~Fan, and P.~Andras, ``{Federated learning with hierarchical
  clustering of local updates to improve training on non-IID data},'' in
  \emph{2020 International Joint Conference on Neural Networks (IJCNN)}.\hskip
  1em plus 0.5em minus 0.4em\relax IEEE, Sep. 2020.

\end{thebibliography}

\end{document}